1# BA-LINS: A Frame-to-Frame Bundle Adjustment for LiDAR-Inertial Navigation

Hailiang Tang, Tisheng Zhang, Liqiang Wang, Man Yuan, and Xiaoji Niu*Abstract*—Bundle Adjustment (BA) has been proven to improve the accuracy of the LiDAR mapping. However, the BA method has not yet been properly employed in a dead-reckoning navigation system. In this paper, we present a frame-to-frame (F2F) BA for LiDAR-inertial navigation, named BA-LINS. Based on the direct F2F point-cloud association, the same-plane points are associated among the LiDAR keyframes. Hence, the F2F plane-point BA measurement can be constructed using the same-plane points. The LiDAR BA and the inertial measurement unit (IMU)-preintegration measurements are tightly integrated under the framework of factor graph optimization. An effective adaptive covariance estimation algorithm for LiDAR BA measurements is proposed to further improve the accuracy. We conduct exhaustive real-world experiments on public and private datasets to examine the proposed BA-LINS. The results demonstrate that BA-LINS yields superior accuracy to state-of-the-art methods. Compared to the baseline system FF-LINS, the absolute translation accuracy and state-estimation efficiency of BA-LINS are improved by 29.5% and 28.7% on the private dataset, respectively. Besides, the ablation experiment results exhibit that the proposed adaptive covariance estimation algorithm can notably improve the accuracy and robustness of BA-LINS.

*Index Terms*—Bundle Adjustment, LiDAR-inertial navigation, factor graph optimization, multi-sensor fusion navigation.## Nomenclature

| | |
|---|---|
| $\mathbf{q}, \mathbf{R}, \boldsymbol{\phi}$ | The attitude quaternion, rotation matrix, and rotation vector. |
| $\otimes$ | The quaternion product. |
| Log, Exp | The transformation between the quaternion and rotation vector. |
| $\mathbf{p}$ | A three-dimension position. |
| $\mathbf{n}, d$ | The plane parameters. |
| $\Gamma$ | The plane thickness. |
| $\boldsymbol{\Sigma}$ | The covariance matrix. |
| $\mathbf{p}_{\mathrm{wb}}^{\mathrm{w}}, \mathbf{q}_{\mathrm{b}}^{\mathrm{w}}$ | The IMU pose w.r.t the world frame. |
| $\mathbf{v}_{\mathrm{wb}}^{\mathrm{w}}$ | The IMU velocity in the world frame. |
| $\mathbf{b}_g, \mathbf{b}_a$ | The gyroscope and accelerometer biases. |
| $\mathbf{p}_{\mathrm{br}}^{\mathrm{b}}, \mathbf{q}_{\mathrm{r}}^{\mathrm{b}}$ | The LiDAR-IMU extrinsic parameters. |
| $t_d$ | The time-delay parameter between the LiDAR and the IMU data. |
| $\mathbf{X}, \mathbf{x}$ | The state vector. |

This research is partly funded by the National Natural Science Foundation of China (No.42374034 and No.41974024). (*Corresponding authors: Tisheng Zhang; Xiaoji Niu.*)

Hailiang Tang, Liqiang Wang, and Man Yuan are with the GNSS Research Center, Wuhan University, Wuhan 430079, China (e-mail: thl@whu.edu.cn; wlq@whu.edu.cn; yuanman@whu.edu.cn).

Tisheng Zhang and Xiaoji Niu are with the GNSS Research Center, Wuhan University, Wuhan 430079, China, and also with the Hubei Luojia Laboratory, Wuhan 430079, China (e-mail: zts@whu.edu.cn; xjniu@whu.edu.cn).## I. Introduction

LIGHT detection and ranging (LiDAR)-based navigation has been widely used for autonomous vehicles and robots in recent years, especially with the rapid development of the newest solid-state LiDAR [1]. However, the LiDAR point clouds suffer from motion distortions, which may result in accuracy degradation. The low-cost micro-electro-mechanical system (MEMS) inertial measurement unit (IMU) is usually adopted to correct the motion distortion of the point clouds. Besides, the inertial navigation system (INS) has the capability of autonomous continuous navigation and space-time transfer and thus plays an essential role in multi-sensor fusion navigation systems [2], [3]. Hence, the LiDAR and the MEMS IMU have been integrated to construct the LiDAR-inertial navigation system (LINS) [4], [5], [6], [7]. Tightly-coupled LINSs [5] have been proven more accurate and robust than loosely-coupled LINSs [8] and thus have become mainstream.

Without a prebuilt map, the LINS should be a dead-reckoning (DR) system, and the yaw angle and the 3-dimensional (3D) position may drift over time with growing covariance [9]. In some LINSs, a kind of frame-to-map (F2M) association method for point clouds is adopted, and an absolute measurement model is wrongly constructed [5], [6], [7], [10], [11], leading to inconsistent state estimation [4], [12]. Here, the frame or the scan is a cluster of continuously sampled point clouds from LiDAR. As a consequence, with the F2M-based measurement model, it is impossible to incorporate other absolute-positioning sensors in tightly-coupled forms, such as the ultrawideband (UWB) [13] and the global navigation satellite system (GNSS) [2]. The frame-to-frame (F2F) association methods [4], [14], [15], [16], [17] have been employed to solve the inconsistent problem in state estimation. A relative measurement model can be constructed with the F2F association methods, which satisfies the characteristics of a DR system. Hence, the F2F-based methods can be seamlessly incorporated into a multi-sensor navigation system.

However, the F2F association of LiDAR is challenging to achieve. LIPS [14] segments planes offline from LiDAR frames using the Point Cloud Library (PCL) [18] and associates the planes among multiple frames. LIC-Fusion 2.0 [15] and VILENS [16] extract plane and line points from a LiDAR frame and track them from frame to frame. The F2F

42association methods in [14], [15], [16] need to explicitly extract plane or line points, which may fail in unstructured environments, such as forests and roads without structured objects. Besides, feature extraction and tracking in these methods [14], [15], [16] have high computational complexity, resulting in low efficiency. LIO-Mapping [17] builds a local point-cloud map with the LiDAR frames in the local window and achieves F2F associations between the pivot frame and other frames in the optimization window. However, the pose errors will be introduced into the local map, leading to inaccurate state estimation in subsequent processes. FF-LINS [4] uses the INS pose to accumulate several LiDAR frames to build keyframe point-cloud maps and achieves F2F associations between the latest keyframe and other keyframes in the sliding window. Due to the high short-term accuracy of INS [19], the keyframe point-cloud map is almost unaffected by pose errors in FF-LINS. Nevertheless, previous methods [14], [15], [16] construct continuous F2F associations across several frames, while LIO-Mapping [17] and FF-LINS [4] only construct associations between one frame and another frame.

In terms of form, LIPS [14], LIC-Fusion 2.0 [15], and VILENS [16] are very like the bundle adjustment (BA) in visual multiple view geometry [20]. In LiDAR mapping, BA is first introduced by BALM [21] and has been proven more accurate. In BALM, plane and line points are associated with a plane-point voxel map and a line-point voxel map, respectively. The local BA in BALM is conducted in a sliding window with marginalization information. However, the marginalized points are retained in the voxel map, which is equivalent to a kind of F2M association [21]. Thus, the inconsistent state estimation problem exists in BALM. Hence, BALM is more suitable for LiDAR mapping rather than multi-sensor fusion navigation. Aiming at multi-sensor fusion navigation applications, the F2F BA should be adopted to maintain consistency in state estimation.

In this study, we propose an F2F BA for LiDAR-inertial navigation, named BA-LINS. We first associate the same-plane points among the LiDAR keyframes to incorporate the LiDAR BA method. Then, the LiDAR plane-point BA measurement is constructed by minimizing the plane thickness. Finally, the LiDAR BA and IMU-preintegration measurements are tightly coupled within the factor graph optimization (FGO) framework, with adaptive covariance estimation for LiDAR measurements. The main contributions of this study are as follows:

- We present a consistent LiDAR-inertial navigation system that tightly integrates LiDAR BA and IMU-preintegration measurements using the FGO. The LiDAR-IMU spatiotemporal parameters are calibrated and compensated online.
- An F2F LiDAR BA measurement model is constructed by minimizing the plane thickness of the same-plane points. The LiDAR BA measurement model achieves a multi-state relative pose constraint. The LiDAR BA measurement residuals and the Jacobians for the IMU poses and the LiDAR-IMU extrinsic parameters are all analytically expressed.
- The same-plane points are associated among the LiDAR keyframes based on the direct F2F point-cloud association. Due to this unique association method, an adaptive covariance estimation algorithm for LiDAR BA measurements is presented to improve navigation accuracy.
- Comprehensive experiments on the public and private datasets are conducted to evaluate the proposed BA-LINS. Sufficient experiment results exhibit that the proposed method is more accurate and efficient than the baseline system.

The proposed BA-LINS is built upon our previous work, FF-LINS [4], but it further incorporates the LiDAR BA measurements. The current implementations are mainly designed for solid-state LiDARs with a non-repetitive scanning pattern. The proposed methods should be applicable for spinning LiDARs, but additional work may be necessary to adapt to them.

The remainder of this paper is organized as follows. We give a brief literature review in Section II. The system pipeline of the proposed BA-LINS is provided in Section III. The methodology of this study, including the same-plane point association method and the plane-point BA measurement model, is presented in Section IV. Experiments and results for quantitative evaluation are discussed in Section V. Finally, we conclude the proposed BA-LINS.

## II. RELATED WORKS

This section discusses the related works on LiDAR-inertial odometry (LIO) and LINS. According to the form of the LiDAR measurement model, we classify them into two categories, *i.e.* non-BA methods and BA-like methods. Point-to-plane and point-to-line distance measurement models are usually employed in non-BA methods, such as LIO-SAM [5] and FAST-LIO [6], [7]. In contrast, a multi-state constraint measurement model is adopted in BA-like methods, such as LIC-Fusion 2.0 [15] and VILENS [16]. It should be noted that filtering-based methods are also included in BA-like methods, though BA is typically an optimization-based method.

### A. Non-BA Methods

In LOAM [8], the orientation and acceleration from a 9-axis IMU are utilized to compensate for the motion distortion of point clouds, yielding improved accuracy to the LiDAR-only odometry. The prior pose from IMU is employed to assist the LiDAR odometry in Cartographer [22], which is based on the probability grid map. LIO-SAM [5] integrates the LiDAR odometry and the IMU preintegration to achieve a LiDAR-inertial state estimation under the framework of FGO. The GNSS and loop closure are further incorporated into LIO-SAM to build a simultaneous localization and mapping (SLAM) system. D-LIOM [23] is a similar system that integrates the LiDAR odometry and IMU preintegration. However, these LINSs are all loosely-coupled systems, as the LiDAR odometry is adopted in the state estimator rather than the LiDAR raw measurements.

Tightly-coupled LINSs have been proven to be more accurate than the loosely-coupled LINSs. An iterated extended Kalman filter (IEKF) [24] is designed to ensure both accuracy and efficiency for tightly-coupled LiDAR-inertial navigation [10]. In [10], the extracted plane and line features [10] are matched with global feature maps to achieve state estimation. FAST-LIO [6] extends the work in [10] by using a new formula for computing the Kalman gain in IEKF, exhibiting higher computational efficiency. A global incremental k-d tree is adopted in FAST-LIO2 with direct point-cloud registration [7]. FAST-LIO2 yields improved accuracy and efficiency compared to state-of-the-art (SOTA) systems. Furthermore, Faster-LIO employs a global incremental voxel as the point-cloud spatial data structure, yielding significantly improved efficiency [11]. LiLi-OM [25] proposes a tightly-coupled LINS using a sliding-window optimizer and designs a new feature-extraction algorithm for a new solid-state LiDAR, *i.e.* Livox Horizon. However, the LINSs mentioned above all use the F2M association methods and construct a wrongly absolute measurement model, leading to inconsistent state estimation.

To adequately address the state covariance and achieve consistent DR navigation, the F2F association should be employed in LINS. In LIC-Fusion [26], the F2F association method in LOAM [8] is used to build the LiDAR measurement model but exhibits poor navigation accuracy. In LIO-Mapping [17], a local point-cloud map is built using the LiDAR frames in the local window, and the F2F point-to-plane and point-to-line associations are achieved between the pivot frame and other frames in the optimization window. As the pose states, which are still estimating, are employed to build the local map, their errors may be unavoidably introduced in LIO-Mapping [17], resulting in inaccurate state estimation. In contrast, the short-term accuracy of the INS is used in FF-LINS [4], and keyframe point-cloud maps are built with only several LiDAR frames using the INS prior pose. With the keyframe point-cloud maps, FF-LINS achieves F2F associations between the latest keyframe and other keyframes in the sliding window. However, due to the direct point-cloud processing without feature extraction, FF-LINS [4] has to set a large standard deviation (STD) for the F2F point-to-plane measurements, *i.e.* 0.1 m, to maintain the robustness. As 0.1 m is larger than the measurement noise of a normal LiDAR, *e.g.* 0.05 m, the LiDAR accuracy has not yet been entirely performed.

F2F association methods are adopted in the above systems [4], [17], [26], and thus the inconsistent problems should be solved. Nevertheless, the F2F measurement models in [4], [17], [26] are built between one frame and another frame, which are relatively dispersed. The reason is that they failed to associate a kind of same-name points across multiple frames, just like visual features tracking [27]. Hence, these methods can be further improved by achieving the same-name points association and constructing a BA-like measurement model.

*B. BA-Like Methods*

BA has been widely employed in visual-based 3D reconstruction [20] and navigation by using continuously tracked features [28], [29]. Besides, a local visual BA [2], [30], [31] is usually used for real-time navigation. Although the data association of the LiDAR point clouds is a challenging task, some work has been conducted to achieve a BA-like LiDAR navigation or mapping [14], [15], [16], [21], [32], [33], [34].

LIPS [14] uses random sample consensus (RANSAC) [24] plane segmentation in PCL [18] to find planar subsets. An anchor plane factor is proposed to build a relative measurement model in graph-based optimization [14]. As an anchor plane may be associated with the points from multiple LiDAR frames, LIPS is a BA-like method. However, the plane segmentation in LIPS [14] should be conducted offline, exhibiting poor computational efficiency. The extracted plane point features [8] are associated by a normal-based method in LIC-Fusion 2.0 [15], and the tracked plane features are divided into multi-state constraint Kalman filter (MSCKF)-based [35], [36], [37] and SLAM-based plane landmarks. Thus, the LiDAR plane measurements are converted into a BA-like form. Similarly, VILENS [16] extracts plane and line features from a LiDAR frame and tracks them from frame to frame. LiDAR plane and line factors are constructed with the tracked features in a BA-like sliding-window factor graph structure [16]. Nevertheless, feature-based methods [15], [16] are mainly designed for structured environments, as it is hard to extract plane or line features in unstructured environments, resulting in poor robustness. Besides, feature extraction and tracking may cost enormous computational resources; thus, these methods are unsuitable for real-time navigation in complex environments.

BALM [21] proposes a kind of plane-point and line-point BA for LiDAR mapping, yielding improved accuracy. The extracted feature points are associated with adaptive voxel maps, and a local BA is conducted by minimizing the eigenvalues of the covariance matrix of the points within a voxel. However, global feature voxel maps must be employed for data association in BALM, which may significantly increase memory costs, especially in large-scale environments. Hence, BALM [21] is more suitable for LiDAR mapping than real-time navigation. Besides, the marginalized points are retained in the voxel, equivalent to the F2M association, leading to inconsistent state estimation. The works in BALM are extended to offline LiDAR mapping in BALM 2.0 [32]. In the meantime, a hierarchical LiDAR BA method is proposed for large-scale LiDAR consistent mapping [33]. Recently, the BA method in [21] is incorporated into a LiDAR-inertial system for back-end map refining, *i.e.* BA-LIOM [34], yielding superior robustness and accuracy. However, the BA method has only been employed for LiDAR-only mapping rather than tightly-coupled LiDAR-inertial navigation.

In conclusion, the F2F BA method for consistent LiDAR-inertial navigation in a tightly-coupled form has not yet been studied. Hence, we propose BA-LINS, an F2F BA method for tightly-coupled LiDAR-inertial navigation, so as to improve navigation accuracy. An effective same-plane point association method is presented to achieve a multi-frame LiDAR data association. In the meantime, an F2F plane-point BA measurement model is proposed to incorporate the LiDAR BA method under the framework of FGO. In the meantime, an adaptive covariance estimation algorithm for LiDAR BA measurements is presented to fully utilize the accurate LiDAR measurements.

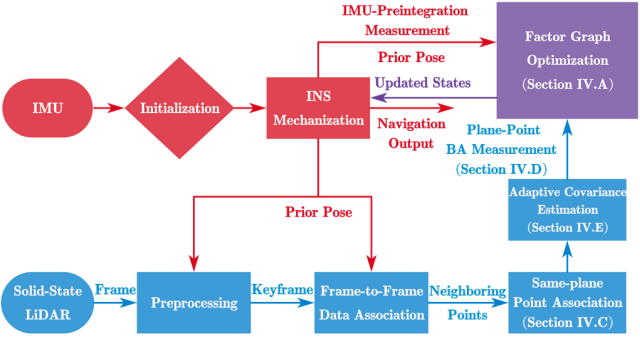

Fig. 1. System pipeline of the proposed BA-LINS. The proposed methods are presented in Section IV. We adopt a direct point-cloud preprocessing method without explicit feature extraction, and we treat all point clouds as plane-point candidates. Hence, only plane-point BA is achieved in BA-LINS.

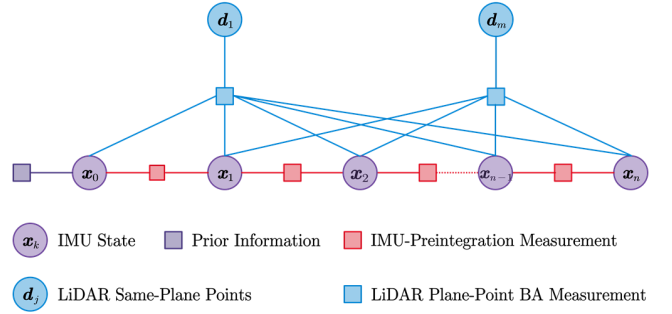

Fig. 2. FGO framework of the proposed BA-LINS. The LiDAR plane-point BA measurement model constructs a multi-state constraint.

## III. SYSTEM OVERVIEW

The system pipeline of the proposed BA-LINS is depicted in Fig. 1. We follow our previous work to adopt an INS-centric processing structure [4]. Once the INS is initialized, the INS mechanization is conducted to output the continuous prior pose. Here, a simple static initialization is employed to obtain the roll and pitch angles and the gyroscope biases [2]. The high-frequency INS pose is adopted to assist the direct LiDAR preprocessing, including motion distortion correction and keyframe selection. We do not explicitly extract plane or line feature points from a LiDAR frame while treating all LiDAR points as plane-point candidates. Besides, we build a keyframe point-cloud map with all LiDAR frames since the last keyframe for further data association.

With the keyframe point-cloud maps, we can build F2F data associations between the latest LiDAR keyframe and other keyframes in the sliding window by finding neighboring points in each keyframe point-cloud map. Then, same-plane point associations can be conducted using the nearest neighboring points from the successful F2F associations. We can also derive adaptively estimated covariances for plane-point BA measurements during the same-plane point association. Finally, the LiDAR plane-point BA and IMU-preintegration measurements are tightly fused under the framework of FGO to perform a maximum-a-posterior estimation.

## IV. METHODOLOGY

This section presents the methodology of the proposed BA-LINS. Firstly, the problem formulation of the tightly-coupled LiDAR-IMU state estimation is provided. Next, we define the plane thickness to derive the LiDAR BA measurement model. Then, we present the F2F same-plane point association method to obtain the plane points across multiple frames. Finally, we derived the analytical form of the LiDAR plane-point BA measurement model with an adaptive covariance estimation method.

### A. Problem Formulation

The proposed tightly-coupled LiDAR-inertial navigation state estimator is a sliding-window optimizer, which balances the accuracy and efficiency. The LiDAR F2F BA and IMU-preintegration measurements are all relative constraints; thus, the proposed BA-LINS is consistent in state estimation. Fig. 2 exhibits the FGO framework of the proposed BA-LINS. The LiDAR plane-point BA measurement model constructs a multi-state constraint across multiple IMU states, and the IMU-preintegration measurement builds a relative constraint for two consecutive IMU states.

The state vector $X$ in the sliding window is defined as follows

$$\boldsymbol{x}_k = \left[\boldsymbol{p}^{\mathrm{w}}_{\mathrm{wb}_k}, \boldsymbol{q}^{\mathrm{w}}_{\mathrm{b}_k}, \boldsymbol{v}^{\mathrm{w}}_{\mathrm{wb}_k}, \boldsymbol{b}_{g_k}, \boldsymbol{b}_{a_k}\right], \boldsymbol{x}^{\mathrm{b}}_{\mathrm{r}} = \left[\boldsymbol{p}^{\mathrm{b}}_{\mathrm{br}}, \boldsymbol{q}^{\mathrm{b}}_{\mathrm{r}}\right], \quad (1)$$
$$\boldsymbol{X} = \left[\boldsymbol{x}_0, \boldsymbol{x}_1, ..., \boldsymbol{x}_n, \boldsymbol{x}^{\mathrm{b}}_{\mathrm{r}}, t_d\right],$$

where $\boldsymbol{x}_k$ is the IMU state at each time node, and we have $k \in [0, n]$; $n$ is the sliding-window size; the IMU state includes the position $\boldsymbol{p}^{\mathrm{w}}_{\mathrm{wb}_k}$, the attitude quaternion $\boldsymbol{q}^{\mathrm{w}}_{\mathrm{b}_k}$ [38], and the velocity $\boldsymbol{v}^{\mathrm{w}}_{\mathrm{wb}_k}$ of the IMU frame (b-frame) w.r.t the world frame (w-frame), and the gyroscope biases $\boldsymbol{b}_g$ and the accelerometer biases $\boldsymbol{b}_a$; $\boldsymbol{x}^{\mathrm{b}}_{\mathrm{r}}$ is the LiDAR-IMU extrinsic parameters, where r denotes the LiDAR frame (r-frame); $t_d$ represents the time-delay parameter between the LiDAR and the IMU data. Here, the w-frame is defined at the initial point with zero position and zero yaw angle, while the roll and pitch angles are gravity-aligned [2].

The FGO problem in Fig. 2 can be solved by minimizing the sum of the Mahalanobis norm of the LiDAR BA and IMU-preintegration measurements and the prior information as

$$\arg\min_{\boldsymbol{X}} \frac{1}{2} \left\{ \sum_{j \in [1,m]} \left\| \mathbf{r}_R\left(\tilde{\boldsymbol{z}}^R_j, \boldsymbol{X}\right) \right\|^2_{\boldsymbol{\Sigma}^R_j} + \sum_{k \in [1,n]} \left\| \mathbf{r}_{Pre}\left(\tilde{\boldsymbol{z}}^{Pre}_{k-1,k}, \boldsymbol{X}\right) \right\|^2_{\boldsymbol{\Sigma}^{Pre}_{k-1,k}} + \left\| \mathbf{r}_p - \mathbf{H}_p \boldsymbol{X} \right\|^2 \right\}, \quad (2)$$

where $\mathbf{r}_R$ are the residuals for the LiDAR plane-point BA

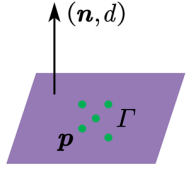

Fig. 3. An illustration of the concept of the plane thickness.

measurements, and $m$ denotes the number of the LiDAR BA measurements; $\Sigma^R$ denotes the covariance matrix of the LiDAR measurement; $\mathbf{r}_{Pre}$ denotes the residuals for the IMU-preintegration measurements, and $\Sigma^{Pre}$ is its covariance matrix; $\{\mathbf{r}_p, \mathbf{H}_p\}$ represent the prior information from the marginalization, and we refer to [39], [40] for more details about the marginalization. We follow our previous work [41] to incorporate the IMU preintegration. The residuals of the IMU preintegration measurements can be written as

$$e^{Pre}\left(\tilde{\mathbf{z}}_{k-1,k}^{Pre}, \mathbf{X}\right) = \begin{bmatrix} \left(\mathbf{R}_{b_{k-1}}^{w}\right)^T \begin{pmatrix} \mathbf{p}_{wb_k}^{w} - \mathbf{p}_{wb_{k-1}}^{w} - \mathbf{v}_{wb_{k-1}}^{w}\Delta t_{k-1,k} \\ -\frac{1}{2}\mathbf{g}^{w}\Delta t_{k-1,k}^2 \end{pmatrix} - \Delta \tilde{\mathbf{p}}_{k-1,k}^{Pre} \\ \left(\mathbf{R}_{b_{k-1}}^{w}\right)^T \left(\mathbf{v}_{wb_k}^{w} - \mathbf{v}_{wb_{k-1}}^{w} - \mathbf{g}^{w}\Delta t_{k-1,k}\right) - \Delta \tilde{\mathbf{v}}_{k-1,k}^{Pre} \\ \mathrm{Log}\left(\left(\mathbf{q}_{b_k}^{w}\right)^{-1} \otimes \mathbf{q}_{b_{k-1}}^{w} \otimes \tilde{\mathbf{q}}_{k-1,k}^{Pre}\right) \\ \mathbf{b}_{g_k} - \mathbf{b}_{g_{k-1}} \\ \mathbf{b}_{a_k} - \mathbf{b}_{a_{k-1}} \end{bmatrix}, \quad (3)$$

where $\Delta \tilde{\mathbf{p}}_{k-1,k}^{Pre}$, $\Delta \tilde{\mathbf{v}}_{k-1,k}^{Pre}$, and $\tilde{\mathbf{q}}_{k-1,k}^{Pre}$ are the position, velocity, and attitude preintegration measurements [41], respectively; $\mathbf{R}_b^w$ denotes the rotation matrix of the quaternion $\mathbf{q}_b^w$; $\mathbf{g}^w$ denotes the gravity vector in the w-frame; $\mathrm{Log}(\bullet)$ represents the transformation from the quaternion to the rotation vector [38]; $\Delta t_{k-1,k}$ is the time length between the two IMU states. The covariance matrix $\Sigma^{Pre}$ is obtained by noise propagation [41].

The Levenberg-Marquardt algorithm [24] in Ceres solver [42] is adopted to solve the non-linear least squares problem in (2). We also employ the two-step optimization in FF-LINS [4] for outlier culling to improve the robustness. Besides, the Huber robust cost function [42] is used for LiDAR measurements to reduce the impact of the outliers.

*B. Definition of the Plane Thickness*

We first define the concept of the plane thickness to derive the LiDAR plane-point measurement model. The plane equation can be written as

$$\mathbf{n}^T \mathbf{p} + d = 0, \quad (4)$$

where $\mathbf{n}$ is the normalized normal vector of the plane; $\mathbf{p}$ is a point on the plane; $d$ is a distance that satisfies the equation. For a cluster of points, *e.g.* five points, an overdetermined linear equation can be built using (4) to solve the plane parameters $(\mathbf{n}, d)$, as depicted in Fig. 3. The point-to-plane

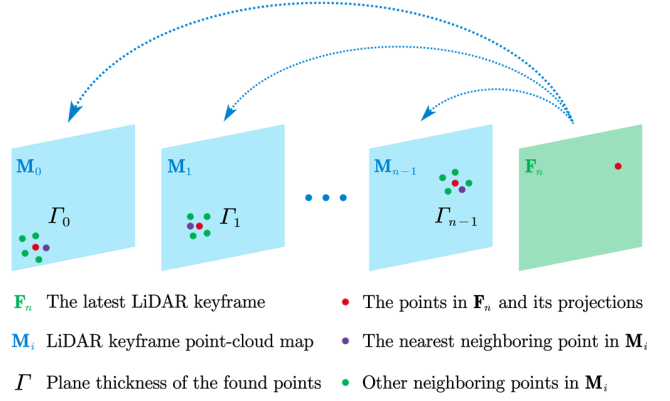

Fig. 4. The same-plane point association and the plane thickness of the neighboring points. We find five points in each keyframe point-cloud map.

distance $\varepsilon$ for a point $\mathbf{p}$ to the fitted plane $(\mathbf{n}, d)$ is written as

$$\varepsilon = \mathbf{n}^T \mathbf{p} + d. \quad (5)$$

Hence, the plane thickness $\Gamma$ can be defined as the average square point-to-plane distance for all points

$$\Gamma = \frac{1}{N}\sum_{i=1}^{N}(\varepsilon_i)^2 = \frac{1}{N}\sum_{i=1}^{N}\left(\mathbf{n}^T \mathbf{p}_i + d\right)^2, \quad (6)$$

where $N$ is the point number. The $\Gamma$ reflects the points distribution concerning the plane, *i.e.* the plane thickness. Suppose that the measurement noise of the point-to-plane distance $\varepsilon$ for all the points satisfies an independent zero-mean Gaussian distribution $\mathcal{N}\left(0, \sigma_\varepsilon^2\right)$, and $\sigma_\varepsilon$ is the STD. According to the property of random variables, the covariance of the plane thickness $\Sigma^\Gamma$ satisfies the following equation

$$\Sigma^\Gamma = \sigma_\Gamma^2 = 2\left(\sigma_\varepsilon^2\right)^2, \quad (7)$$

where $\sigma_\Gamma$ is the STD of the plane thickness measurement. The equation (7) is helpful for further same-plane point association.

*C. Same-Plane Point Association*

The F2F same-plane point association should be achieved to build the plane-point measurement model. Fig. 4 depicts an illustration of the proposed same-plane point association method. The F2F data association is first conducted by projecting the points in the latest keyframe into other keyframe point-cloud maps and finding five neighboring points, as proposed in FF-LINS [4]. The five neighboring points are employed to fit a plane, and the point-to-plane distance is checked to validate the association. For a successful F2F data association, we obtain five points corresponding to the keyframe, as the green and purple points in Fig. 4.

We finally build successful F2F associations in multiple keyframes for an original point in the latest keyframe. As the F2F association is based on a plane assumption, we have reasons to believe that the found neighboring points and the original point belong to a common physical plane. Hence, we can use these points to construct the plane-point BA measurement model. Specifically, we only pick up the nearest neighboring points and the original point as the same-plane point observations to bound the computational complexity. As shown in Fig. 4, the purple point in each keyframe and the red point in the latest keyframe are the same-plane point candidates. The ablation experiment will be conducted to evaluate the impact of the same-plane point selection in Section V.E. Note that each keyframe point-cloud map is downsampled by a voxel-grid filter [18]. Thus, the selected same-plane points are also dispersed in space.

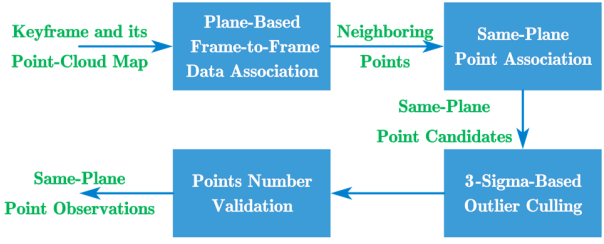

Fig. 5. An overview of the same-plane point association method.

We also adopt an outlier culling algorithm to detect and remove outlier points. The associated same-plane points $\boldsymbol{p}^{r_i}, i \in \mathbb{C}$ are all in the LiDAR frames corresponding to each keyframe, where $\mathbb{C}$ represents the keyframe collections of the successful associations. We project all the points into the w-frame as

$$\begin{aligned}\boldsymbol{p}^{b_i} &= \mathbf{R}_r^b \boldsymbol{p}^{r_i} + \boldsymbol{p}_{br}^b, \\ \boldsymbol{p}^{w_i} &= \mathbf{R}_{b_i}^w \boldsymbol{p}^{b_i} + \boldsymbol{p}_{wb_i}^w,\end{aligned} \quad (8)$$

where $\{\boldsymbol{p}_{br}^b, \mathbf{R}_r^b\}$ is the LiDAR-IMU extrinsic parameters in (1); $\{\boldsymbol{p}_{wb_i}^w, \mathbf{R}_{b_i}^w\}$ is the IMU pose state; $\boldsymbol{p}^{b_i}$ and $\boldsymbol{p}^{w_i}$ are the projections in the b-frame and w-frame, respectively. Hence, we obtain a cluster of points $\boldsymbol{p}^{w_i}, i \in \mathbb{C}$ in the w-frame, and we fit a plane using these points. The point-to-plane distance $\varepsilon^{w_i}$ for $\boldsymbol{p}^{w_i}$ can be then calculated. Assuming we have the covariance of the plane thickness $\Sigma^\Gamma$, we can derive the STD for the point-to-plane distance measurement $\sigma_\varepsilon^w$ using (7). The covariance-estimation method will be presented in Section IV.E. For those points whose point-to-plane distance $\varepsilon^{w_i}$ is not within $\pm 3\sigma_\varepsilon^w$, we treat them as outliers.

However, a plane can only be determined by at least three points on it. Hence, we need more than three same-plane points for a valid BA constraint. Specifically, only those same-plane point associations with at least five points will be treated as successful associations. Finally, we obtain a cluster of same-plane point observations corresponding to multiple keyframes with outliers removed. Fig. 5 depicts an overview of the proposed same-plane point association methods.

### D. Plane-Point BA Measurement Model

With the associated same-plane points, the plane-point BA measurement model can be formulated by minimizing the thickness of the plane constructed by these points. Fig. 6 illustrates the proposed LiDAR plane-point BA model, which forms a multi-state constraint. Assume we have a cluster of same-plane point observations $\tilde{\boldsymbol{p}}^{r_i}, i \in \mathbb{C}_j$, where $\mathbb{C}_j$ represents the keyframe collections of the same-plane point association $j$, and the keyframe number in $\mathbb{C}_j$ is $N$. Then, we can derive the analytical form of the residuals and Jacobians for the proposed BA measurement model. The time-delay parameter $t_d$ between the LiDAR and IMU will be omitted in the following part for convenience, and we refer to [43] for more details.

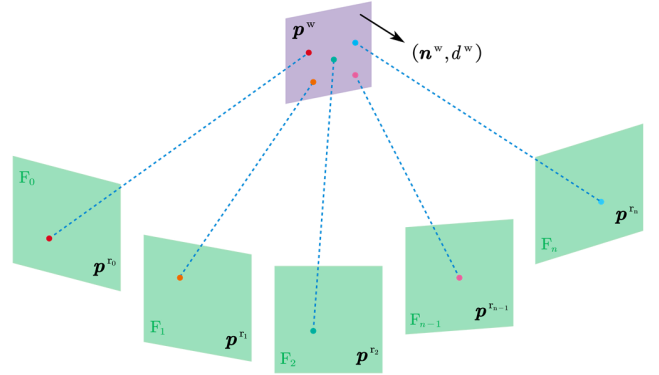

Fig. 6. An illustration of the LiDAR plane-point BA. Points $\boldsymbol{p}^r$ with different colors denote the points in different LiDAR keyframes $\mathbf{F}$. The LiDAR plane-point BA measurement model imposes a multi-state constraint, which is very like the visual BA.

*1) Residuals of the Plane-Point BA Measurement*

The residuals of the proposed plane-point BA measurement are equivalent to the plane thickness. As depicted in Fig. 6, the same-plane points are projected to the w-frame to obtain the residuals. We can also transform the points to a local r-frame, but the w-frame is the most convenient choice. With the projected points $\boldsymbol{p}^{w_i}$ from (8), the plane parameters can be obtained as $(\boldsymbol{n}^w, d^w)$. Finally, the residuals for the plane-point BA measurement can be written as

$$\mathbf{r}_R\left(\tilde{\boldsymbol{z}}_j^R, \boldsymbol{X}\right) = \frac{1}{N} \sum_{i \in \mathbb{C}_j}^{N} \left(\left(\boldsymbol{n}^w\right)^T \boldsymbol{p}^{w_i} + d^w\right)^2. \quad (9)$$

The more specific formulation is calculated by bringing (8) into (9) as

$$\begin{aligned}&\mathbf{r}_R\left(\tilde{\boldsymbol{z}}_j^R, \boldsymbol{X}\right) = \\ &\frac{1}{N} \sum_{i \in \mathbb{C}_j}^{N} \left(\left(\boldsymbol{n}^w\right)^T \left(\mathbf{R}_{b_i}^w \left(\mathbf{R}_r^b \tilde{\boldsymbol{p}}^{r_i} + \boldsymbol{p}_{br}^b\right) + \boldsymbol{p}_{wb_i}^w\right) + d^w\right)^2.\end{aligned} \quad (10)$$

(10) defines the residuals for the proposed plane-point BA measurement model, and its covariance matrix will be presented in Section IV.E.

*2) Jacobians of the Plane-Point BA Measurement*

The plane-point BA measurement residuals in (10) are the function of the pose states $\{\boldsymbol{p}_{wb_i}^w, \mathbf{q}_{b_i}^w\}, i \in \mathbb{C}_j$ and the LiDAR-IMU extrinsic parameters $\{\boldsymbol{p}_{br}^b, \mathbf{q}_r^b\}$. Hence, we can derive the analytical Jacobians of $\mathbf{r}_R$ w.r.t the IMU pose errors $\{\delta \boldsymbol{p}_{wb_i}^w, \delta \boldsymbol{\phi}_{wb_i}^w\}$ and the LiDAR-IMU extrinsic errors $\{\delta \boldsymbol{p}_{br}^b, \delta \boldsymbol{\phi}_r^b\}$, using the error-perturbation method [3]. Here, $\phi$ represents the rotation vector of a quaternion $\mathbf{q}$, and $\delta\phi$ denotes the attitude errors. Specifically, the Jacobians w.r.t the pose errors $\{\delta \boldsymbol{p}_{\mathrm{wb}_i}^{\mathrm{w}}, \delta\phi_{\mathrm{wb}_i}^{\mathrm{w}}\}$ can be formulated as

$$\begin{cases} \dfrac{\partial \mathbf{r}_R}{\partial \delta \boldsymbol{p}_{\mathrm{wb}_i}^{\mathrm{w}}} = \mathbf{J}_{\boldsymbol{p}^{\mathrm{w}_i}} \\ \dfrac{\partial \mathbf{r}_R}{\partial \delta \phi_{\mathrm{wb}_i}^{\mathrm{w}}} = -\mathbf{J}_{\boldsymbol{p}^{\mathrm{w}_i}} \mathbf{R}_{\mathrm{b}_i}^{\mathrm{w}} \left[ \mathbf{R}_{\mathrm{r}}^{\mathrm{b}} \tilde{\boldsymbol{p}}^{\mathrm{r}_i} + \boldsymbol{p}_{\mathrm{br}}^{\mathrm{b}} \right]_{\times} \end{cases}, \quad (11)$$

where $[\cdot]_{\times}$ is the skew-symmetric matrix of a 3D vector [38]. The common part $\mathbf{J}_{\boldsymbol{p}^{\mathrm{w}}}$ in (11) can be written as

$$\mathbf{J}_{\boldsymbol{p}^{\mathrm{w}_i}} = \frac{2}{N}\Big( \big(\boldsymbol{n}^{\mathrm{w}}\big)^T \big(\mathbf{R}_{\mathrm{b}_i}^{\mathrm{w}} \big(\mathbf{R}_{\mathrm{r}}^{\mathrm{b}} \tilde{\boldsymbol{p}}^{\mathrm{r}_i} + \boldsymbol{p}_{\mathrm{br}}^{\mathrm{b}}\big) + \boldsymbol{p}_{\mathrm{wb}_i}^{\mathrm{w}} \big) + d^{\mathrm{w}} \Big) \big(\boldsymbol{n}^{\mathrm{w}}\big)^T. \quad (12)$$

Similarly, we can obtain the Jacobians w.r.t the LiDAR-IMU extrinsic errors $\{\delta \boldsymbol{p}_{\mathrm{br}}^{\mathrm{b}}, \delta \phi_{\mathrm{r}}^{\mathrm{b}}\}$ as

$$\begin{cases} \dfrac{\partial \mathbf{r}_R}{\partial \delta \boldsymbol{p}_{\mathrm{br}}^{\mathrm{b}}} = \sum_{i \in \mathbb{C}_j}^{N} \mathbf{J}_{\boldsymbol{p}^{\mathrm{w}_i}} \mathbf{R}_{\mathrm{b}_i}^{\mathrm{w}} \\ \dfrac{\partial \mathbf{r}_R}{\partial \delta \phi_{\mathrm{r}}^{\mathrm{b}}} = -\sum_{i \in \mathbb{C}_j}^{N} \mathbf{J}_{\boldsymbol{p}^{\mathrm{w}_i}} \mathbf{R}_{\mathrm{b}_i}^{\mathrm{w}} \mathbf{R}_{\mathrm{r}}^{\mathrm{b}} \left[ \tilde{\boldsymbol{p}}^{\mathrm{r}_i} \right]_{\times} \end{cases}. \quad (13)$$

Finally, we obtain the analytical Jacobians of the residuals in (11) and (13). It should be noted that the errors of the plane parameters $(\boldsymbol{n}^{\mathrm{w}}, d^{\mathrm{w}})$ caused by the IMU pose errors and the LiDAR-IMU extrinsic errors are not considered in (11) and (13). The reason is that the impact of the plane parameters $(\boldsymbol{n}^{\mathrm{w}}, d^{\mathrm{w}})$ is very tiny to the residuals of the plane-point BA measurements. Hence, they are omitted to reduce the computational complexity. Ablation experiments will be conducted to verify the impact of these tiny terms in Section V.F.

### E. Adaptive Covariance Estimation

The proposed F2F plane-point BA measurement model constructs a multi-state constraint for the IMU pose states in the sliding window. Hence, it should be more accurate than the dispersed plane-to-point distance measurement model in FF-LINS [4]. Nevertheless, the measurement covariance should be appropriately addressed to bring the accurate LiDAR measurements into full play.

Thanks to the unique same-plane point association method in Section IV.C, an adaptive covariance estimation algorithm for the proposed plane-point BA measurement model can be employed. Specifically, once a successful F2F data association is built, we obtain five neighboring points in the keyframe point-cloud map corresponding, as depicted in Fig. 4. Then, the thickness of the plane, which consists of these five points, can be calculated by (6). For all the F2F data associations from an original point in the latest keyframe, we obtain a cluster of plane thickness $\Gamma_i$, as shown in Fig. 4. As the same-plane points and these neighboring points are assumed to belong to a same physical plane, the plane thicknesses $\Gamma_i$ can reflect the noise of the plane-point BA measurement model.

Hence, the covariance of the plane-point BA measurement residuals in (10) can be derived by quantitative statistics of the plane thicknesses $\Gamma_{i,j}, i \in \mathbb{C}_j, i \neq n$. Here, $\mathbb{C}_j$ denotes the keyframe collections of the same-plane point association $j$. Finally, the adaptive covariance can be calculated as

$$\boldsymbol{\Sigma}_j^R = \frac{1}{N-1} \sum_{i \in \mathbb{C}_j, i \neq n}^{N-1} \big(\Gamma_{i,j}\big)^2 \mathbf{I}, \quad (14)$$

where $N$ is keyframe number in $\mathbb{C}_j$. As we cannot obtain $\Gamma_n$ in the latest keyframe, the number is $N-1$ in (14). It should be noted that $\Gamma_{i,j}$ are calculated by the points of the local r-frame while the same-plane BA measurement residuals are implemented in the w-frame. Nevertheless, the property of the plane thickness should be the same in both the r-frame and the w-frame. Thus, the formulation in (14) is correct. Ablation experiments will be conducted to verify the importance of the proposed adaptive covariance estimation algorithm.

## V. EXPERIMENTS AND RESULTS

Experiments and results will be presented in this section to examine the proposed BA-LINS. The implementation of BA-LINS and employed public and private datasets are described first. Then, quantitative experiments are conducted to evaluate the accuracy and efficiency improvement compared to the SOTA LINSs. Finally, a series of ablation experiments are presented to verify the possible factors that may affect the accuracy of the proposed BA-LINS.

### A. Implementation and Datasets

The proposed BA-LINS is built upon FF-LINS [4] by incorporating the F2F BA for LiDAR measurements. BA-LINS is implemented using C++ with the robot operation system (ROS) supported. Besides, we also employ multi-threading technology in point-cloud processes, such as distortion removal and data association, to improve the computational efficiency.

The employed public datasets are the *R3LIVE* [44] and

TABLE I
DATASETS DESCRIPTIONS

| Datasets | Descriptions |
|---|---|
| R3LIVE | Livox AVIA (10 Hz). BOSCH BMI088 (200 Hz). 3 sequences with a total length of 3878 m and 3405 s. Quantitative evaluation with end-to-end reference. |
| WHU-Helmet | Livox AVIA (10 Hz). MEMS IMU with a gyroscopy bias-instability of 3 º/hr (600 Hz). 4 sequences with a total length of 3654 m and 6031 s. Quantitative evaluation with ground-truth pose. |
| Robot | Livox Mid-70 (10 Hz). ADI ADIS16465 (200 Hz). 8 sequences with a total length of 15248 m and 11296 s. Quantitative evaluation with ground-truth pose. |

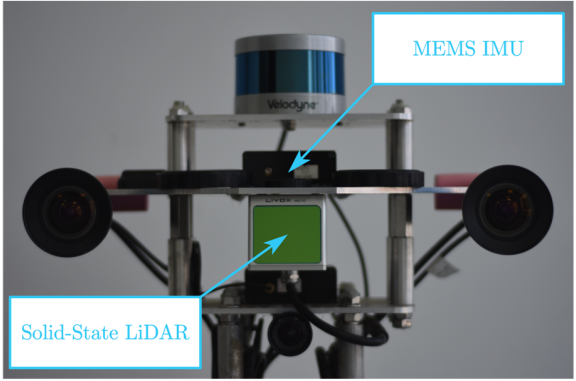

Fig. 7. Equipment setup in the *Robot* dataset.

*WHU-Helemt* [45] datasets. The *R3LIVE* dataset is collected by a handheld device, while the *WHU-Helemt* dataset is collected by a head-mounted device. These datasets are all equipped with a solid-state LiDAR, *i.e.* Livox AVIA, and a low-cost MEMS IMU. The three longest sequences in the *R3LIVE* with the end-to-end reference are adopted for quantitative evaluation, and the total length is about 3878 m. In the *WHU-Helemt* dataset, four sequences with the ground-truth pose are employed, and the whole length is about 3654 m. More details about the public datasets are shown in Table I.

The private *Robot* dataset is collected by a wheeled robot with a maximum speed of 1.5 m/s. The employed sensors include a Livox Mid-70 and a MEMS IMU (ADI ADIS16465), as depicted in Fig. 7 and Table I. The ground-truth pose (0.02 m for position and 0.01 deg for attitude) is generated by post-processing software using a navigation-grade [3] GNSS/INS integrated navigation system. These sensors are well synchronized through hardware triggers. In addition to the four sequences employed in FF-LINS, four large-scale sequences are added. Thus, eight sequences with a total length of 15248 m are included in the *Robot* dataset. Fig. 8 exhibits the testing scenes of the extra four sequences.

The SOTA LINSs, including LIO-SAM (without loop closure) [5], FAST-LIO2 [7], and FF-LINS [4], are adopted for comparisons. LIO-SAM fails on the *Robot* dataset, as it cannot extract enough features for state estimation from the more sparse LiDAR, Livox Mid-70. Livox AVIA has six laser beams, while Livox Mid-70 only has one. We fail to run BALM [21] on these datasets, mainly because the motion distortion is not compensated, as mentioned in [21]. LINSs have been proven to be more accurate than LiDAR-only systems, such as BALM [21] and LOAM [8], and thus we only compare BA-LINS with other SOTA LINSs. In particular, FF-LINS is treated as the baseline system of BA-LINS to derive the quantitative results about the improvements in accuracy and efficiency. The sliding-window size $n$ for FF-LINS and BA-LINS is set to 10 to bound the computational complexity. The LiDAR-IMU extrinsic parameters are assumed to be unknown for all the systems on these datasets. All the systems are run in real-time on a laptop (Intel i7-13700H) under the ROS framework. FF-LINS and BA-LINS read the ROS bags directly to run at full speed to obtain the

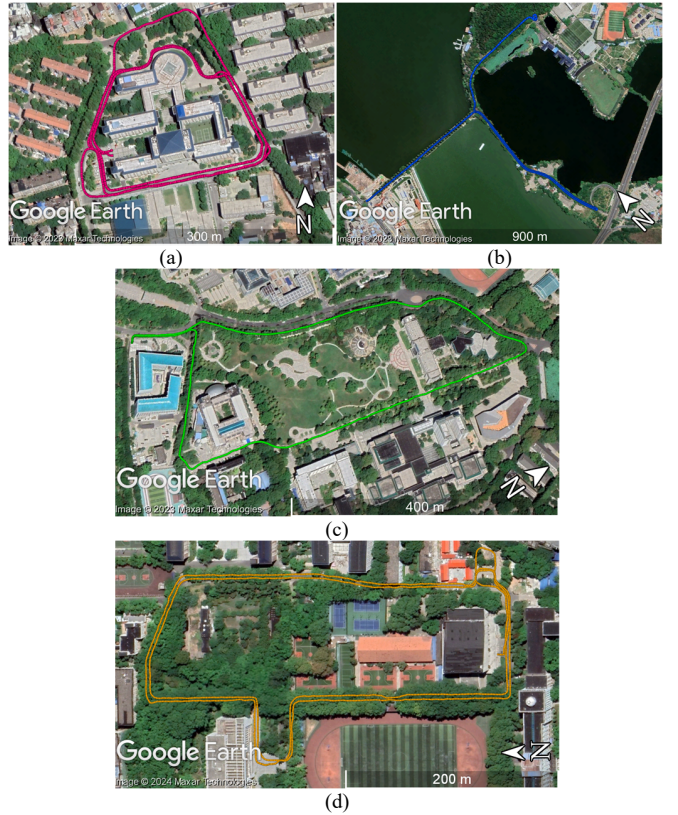

Fig. 8. Testing scenes in the *Robot* dataset. (a) shows the sequence *cs_campus*. (b) shows the sequence *lake_park*. (c) shows the sequence *luojia_square*. (d) shows the sequence *library*. Other testing scenes can be found in FF-LINS.

TABLE II
END-TO-END ERRORS ON THE *R3LIVE* DATASET

| Error (m) | LIO-SAM | FAST-LIO2 | FF-LINS | BA-LINS |
|---|---|---|---|---|
| *hku_main_building* | Failed | 2.50 | 1.20 | **0.65** |
| *hkust_campus_00* | 3.29 | 3.69 | 2.41 | **1.44** |
| *hkust_campus_01* | 20.82 | **0.14** | 2.51 | 2.44 |
| Average | Invalid | 2.11 | 2.04 | **1.51** |

The bold term for each sequence denotes the best result among these LINSs.

statistical results of the efficiency.

### B. Evaluation of the Accuracy
#### 1) Public R3LIVE Dataset

Table II exhibits the end-to-end errors on the *R3LIVE* dataset. LIO-SAM yields the worst accuracy, and it even fails on the *hku_main_building*. The reason is that LIO-SAM cannot extract enough features in the tiny corridor, while other LINSs are all based on the direct method. FAST-LIO2 achieves the minimum error on the *hkust_campus_01*, as it can match its self-built global map using the F2M association method. Due to the consistent F2F state estimator, FF-LINS exhibits improved accuracy than FAST-LIO2 regarding the average error. Furthermore, BA-LINS demonstrates notably improved accuracy compared to FF-LINS on all the sequences. Besides, the average error of BA-LINS is the minimum among these systems, yielding the best accuracy.

The trajectory results on the *hku_main_building* are shown in Fig. 9. At the start-end point, the trajectory of FAST-LIO2 tends to the start point, exhibiting a notable change, as shown in Fig. 9.b. This is because that FAST-LIO2 is trying to match with its previously built point-cloud map. Besides, FAST-LIO2 shows the worst accuracy along the z-axis in Fig. 9.c. In contrast, FF-LINS and BA-LINS exhibit more smooth trajectories and minor errors along the z-axis. Besides, BA-LINS almost returns to zero along the z-axis, yielding improved accuracy than FF-LINS.

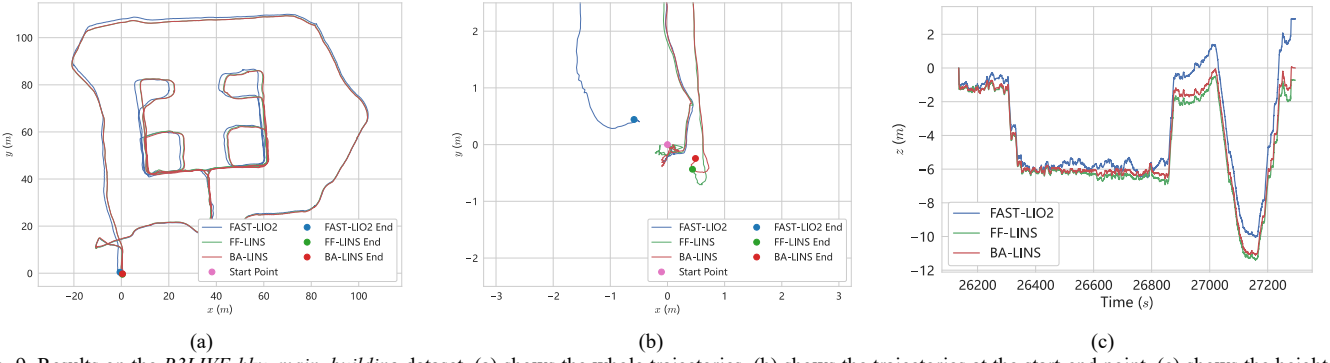

Fig. 9. Results on the *R3LIVE-hku_main_building* dataset. (a) shows the whole trajectories. (b) shows the trajectories at the start-end point. (c) shows the height (z-axis) changes.

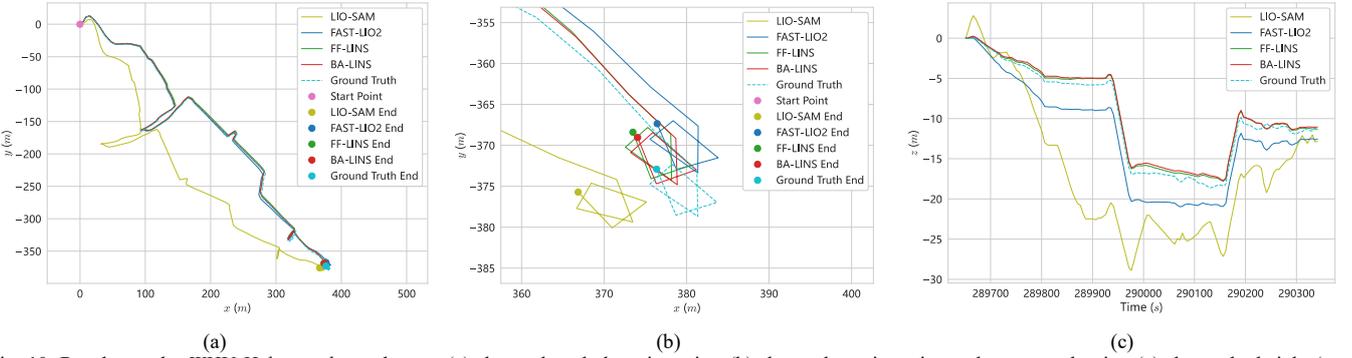

Fig. 10. Results on the *WHU-Helmet-subway* dataset. (a) shows the whole trajectories. (b) shows the trajectories at the start-end point. (c) shows the height (z-axis) changes.

TABLE III
ATEs ON THE *WHU-HELEMT* DATASET

| ATE (m) | LIO-SAM | FAST-LIO2 | FF-LINS | BA-LINS |
|---|---|---|---|---|
| *mall* | 0.55 | **0.32** | 0.69 | 0.55 |
| *residence* | **0.35** | 1.03 | 0.43 | 0.38 |
| *street* | 1.06 | 0.90 | 0.97 | **0.67** |
| *subway* | 28.43 | 2.39 | 2.34 | **1.76** |
| Average | 7.60 | 1.16 | 1.11 | **0.84** |

TABLE IV
AREs AND ATEs ON THE *ROBOT* DATASET

| ARE / ATE (deg / m) | FAST-LIO2 | FF-LINS | BA-LINS |
|---|---|---|---|
| *campus* | 3.55 / 4.42 | 0.41 / 1.51 | **0.40 / 1.24** |
| *building* | 3.13 / 3.12 | 0.65 / 1.90 | **0.57 / 1.82** |
| *playground* | 2.84 / 1.59 | **0.77** / 1.27 | 0.84 / **0.96** |
| *park* | 3.24 / 4.00 | **0.90 / 1.44** | 1.23 / 2.07 |
| *cs_campus* | 3.68 / 4.38 | 0.93 / 2.04 | **0.55 / 1.39** |
| *luojia_square* | 3.47 / 5.18 | 0.88 / 3.88 | **0.54 / 2.72** |
| *east_lake* | 3.20 / 4.49 | 1.48 / 8.39 | **0.85 / 3.57** |
| *library* | 3.28 / 2.73 | **0.37 / 1.77** | 0.49 / 1.90 |
| Average | 3.30 / 3.74 | 0.80 / 2.78 | **0.68 / 1.96** |

*2) Public WHU-Helmet Dataset*

On the *WHU-Helmet* dataset, the absolute translation errors (ATE) [46] are calculated using evo [47], as shown in Table III. LIO-SAM exhibits the worst average accuracy, as it almost fails in the indoor environments of the sequence *subway*. According to the average errors, FAST-LIO2 and FF-LINS almost yield the same accuracy. Compared to FF-LINS, BA-LINS achieves improved accuracy on all four sequences. Besides, BA-LINS achieves the best accuracy on two sequences and the minimum average error. Fig. 10 shows the trajectory results on the sequence *subway*. LIO-SAM exhibits the worst trajectory compared to the ground truth. In contrast, other LINSs show aligned trajectories to the ground truth in Fig. 10.a. BA-LINS and FF-LINS exhibit better-aligned trajectories in Fig. 10.b and Fig. 10.c, especially along the z-axis. Besides, BA-LINS exhibits a more aligned trajectory to the ground truth than FF-LINS, showing improved accuracy.

*3) Private Robot Dataset*

We also evaluate the absolute rotation errors (ARE) and absolute translation errors (ATE) on the *Robot* dataset. We fail to run LIO-SAM on the Robot dataset, as the point clouds of the employed Livox Mid-70 are very sparse to extract enough valid features. Table IV illustrates that FAST-LIOS exhibits the largest average errors, especially for the ARE. The reason is that FAST-LIO2 cannot estimate the LiDAR-IMU extrinsic parameters, mainly the rotation parameters. In contrast, FF-LINS and BA-LINS can achieve online estimation and compensation of the extrinsic parameters. Hence, the AREs for FF-LINS and BA-LINS are much smaller. We can refer to FF-LINS [4] for more details about the impact of the online calibration of the LiDAR-IMU extrinsic parameters. Besides, FF-LINS outperforms FAST-LIO2 regarding the average ARE and ATE due to the consistent F2F state estimator. Furthermore, with the plane-point BA measurement model, BA-LINS exhibits significantly improved accuracy in rotation and translation. More specifically, compared to FF-LINS, the average absolute translation accuracy for BA-LINS is improved by 29.5%.

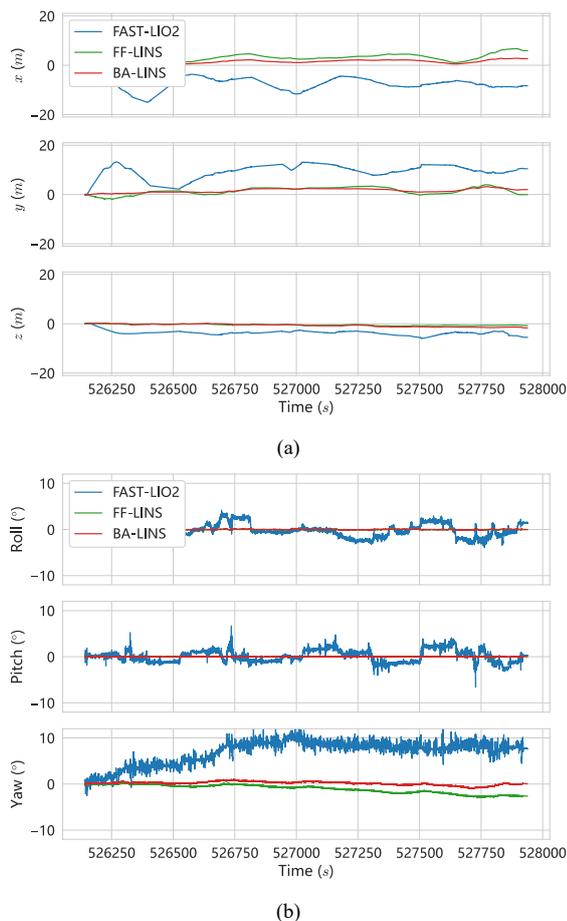

(a)

(b)

Fig. 11. Position and attitude errors on the *Robot-cs_campus* dataset. (a) shows the position errors. (b) shows the attitude errors. Due to the consistent state estimator in FF-LINS and BA-LINS, the roll and pitch angles are observable terms, and thus they may not diverge. In contrast, the yaw angle may diverge due to its unobservability.

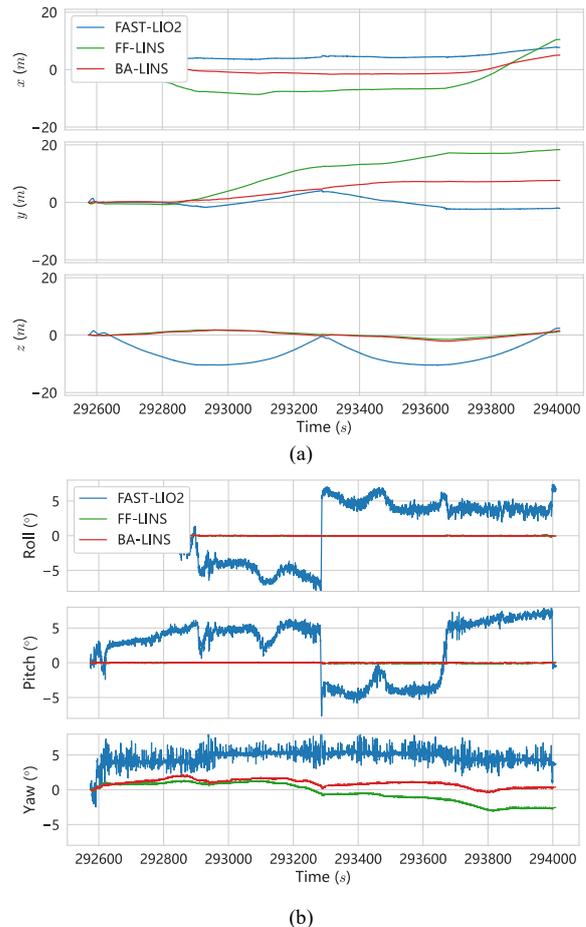

(a)

(b)

Fig. 12. Position and attitude errors on the *Robot-east_lake* dataset. (a) shows the position errors. (b) shows the attitude errors.

We calculate the position and attitude errors along each axis to evaluate the DR capability of BA-LINS. Position and attitude errors on the sequences *cs_campus* and *east_lake* are shown in Fig. 11 and Fig. 12. As for the position error, FAST-LIO exhibits the largest error on the sequence *cs_campus*, while FF-LINS is the worst on the sequence *east_lake*.

Compared to FF-LINS, the proposed BA-LINS shows smaller position erros on both the two sequences, mainly in the horizontal direction, *i.e.* the x and y axes. Regarding attitude errors, FAST-LIO2 shows poor accuracy for the roll, pitch, and yaw angles because the F2M association may result in wrong observability. In contrast, the roll and pitch angles for FF-LINS and BA-LINS are not diverged due to their observability in the consistent F2F state estimator, while the yaw angle may diverge due to its unobservability. By incorporating the accurate plane-point BA measurement model, BA-LINS exhibits superior yaw accuracy than FF-LINS, and yaw errors are less than $1^\circ$ at the end of the sequences, as shown in Fig. 11.b and Fig. 12.b.

### C. Evaluation of the Efficiency

The state-estimation efficiency can also be improved with the proposed plane-point BA measurement model. On the one hand, many F2F point-to-plane associations will be abandoned in the same-plane point association processing. They are mainly outliers or the F2F associations that are less than five same-plane point associations, as depicted in Fig. 5. On the other hand, some computations are not repeated in the plane-point BA measurement model. Specifically, the Jacobians for the latest IMU pose state are calculated multiple times in each point-to-plane measurement in FF-LINS [4]. In contrast, they

11TABLE V
AVERAGE RUNNING TIME OF THE STATE ESTIMATION AND EQUIVALENT FPS
ON THE *ROBOT* DATASET

| \ | Running time (ms) | | Equivalent FPS (Hz) | |
|---|---|---|---|---|
| | FF-LINS | BA-LINS | FF-LINS | BA-LINS |
| *campus* | 27.7 | **19.5** | 48 | **104** |
| *building* | 27.3 | **19.6** | 49 | **105** |
| *playground* | 31.9 | **19.2** | 51 | **113** |
| *park* | 22.8 | **19.8** | 69 | **131** |
| *cs_campus* | 26.6 | **18.6** | 54 | **111** |
| *luojia_square* | 26.8 | **20.6** | 54 | **109** |
| *east_lake* | 24.0 | **18.7** | 58 | **114** |
| *library* | 30.6 | **19.0** | 49 | **110** |
| Average | 27.2 | **19.4** | 54 | **112** |

Here, the equivalent FPS is calculated by dividing the sequence length by the total running time and multiplying by the LiDAR frame rate.

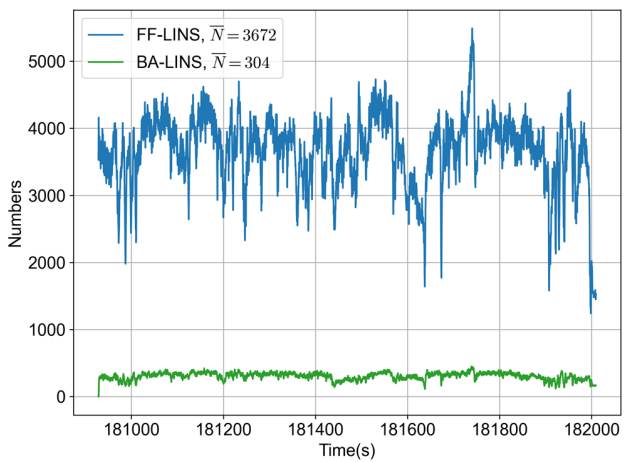

Fig. 13. Comparison of the number of LiDAR measurements on the *Robot-campus* dataset.

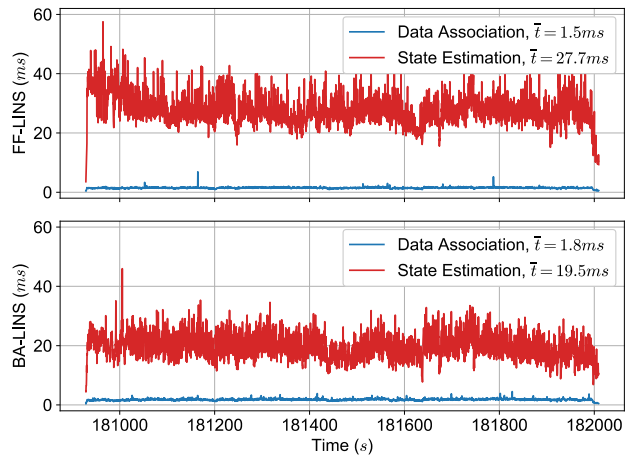

Fig. 14. Comparison of time costs on the *Robot-campus* dataset. The data association for FF-LINS only includes the F2F data association. The data association for BA-LINS includes the F2F data association, the same-plane point association, and the adaptive covariance estimation.

are only calculated once in each plane-point BA measurement. Table V compares the state-estimation running time between FF-LINS and BA-LINS. Quantitative results indicate that the average state-estimation efficiency of BA-LINS is improved by 28.7% compared to the baseline FF-LINS. By further employing multi-threading technology, BA-LINS achieves an average equivalent frame per second (FPS) of 112, twice more than that of FF-LINS. Here, the equivalent FPS is calculated by dividing the sequence length by the total running time and multiplying by the LiDAR frame rate.

Fig. 13 compares the LiDAR measurements number on the *Robot-campus* dataset. The point-to-plane distance measurements are employed in FF-LINS, while the same-plane point measurements are used in the proposed BA-LINS. According to the average measurement number in Fig. 13, we can conclude that many F2F point-to-plane associations are abandoned in BA-LINS. As the sliding-window size is $n = 10$, the number of abandoned associations can be roughly calculated as $3672 - 304 * 10$, *i.e.* 632. That is one of the reasons that the state-estimation efficiency is improved, and the one-time computation for the Jacobians is another reason.

We also derive the statistical results of the time costs for FF-LINS and BA-LINS. Fig. 14 compares the time costs on the *Robot-campus* dataset, including the data association and state estimation. Here, the data association for FF-LINS only includes the F2F data association. In contrast, it consists of the F2F data association, the same-plane point association, and the adaptive covariance estimation for BA-LINS. As depicted in Fig. 14, the average time cost of the data association in BA-LINS only increases by 0.3 ms. In addition, the average time cost of the state estimation decreases by 8.2 ms. Hence, the overall time costs of the proposed BA-LINS are much lower than that of FF-LINS, yielding higher computational efficiency.

*D. The Impact of the Adaptive Covariance Estimation*

An adaptive covariance estimation algorithm is proposed in BA-LINS to fully utilize the accurate LiDAR measurements. Ablation experiments are conducted to evaluate the impact of the proposed adaptive covariance estimation by adopting different covariances. We set the STD of the point-to-plane distance measurement $\sigma_\varepsilon$ as 0.01 m, 0.02 m, and 0.05 m, and $\sigma_\varepsilon$ can be converted to the covariance of the plane-point BA measurement using (7). As shown in Table VI, it fails on two sequences when the $\sigma_\varepsilon$ is 0.01 m. As the STD $\sigma_\varepsilon$ is also employed for outlier culling in the same-plane point association, a minor STD will result in insufficient measurements for state estimation. Nevertheless, the average ARE and ATE are increased notably compared to the proposed method when $\sigma_\varepsilon$ is set as 0.02 m or 0.05 m. The results demonstrate that the proposed adaptive covariance estimation algorithm is effective in improving the navigation accuracy and also the system robustness.

*E. The Impact of the Same-plane Point Selection*

In the proposed same-plane point association method, the nearest point among the five neighboring points is selected as the same-plane point candidate, which is a direct selection. We carry out ablation experiments to verify the impact of the same-plane point selection. We reorder the five neighboring points (the green and purple points in Fig. 4) by the distance relative to the projected point (the red point in Fig. 4). According to the results in Table VI, the average ARE and the average ATE are the minimum when we select the second neighboring point as the same-plane point. The reason may be that the constructed plane by the same-plane points is much larger when the second neighboring point is selected; thus, the plane-point BA measurement model can be more effective. The constructed plane is too large to satisfy the adaptive estimated covariance when the middle or the furthest point is selected. Nevertheless, the ARE and ATE of the proposed BA-LINS are almost the same as the results of selecting the second neighboring point. Hence, the results are acceptable for the proposed BA-LINS.

TABLE VI
AREs AND ATEs ON THE *ROBOT* DATASET WITH DIFFERENT CONFIGURATIONS

| ARE / ATE (deg / m) | Different Covariance[1] | | | Different Selections of Plane Points[2] | | | With the Tiny Terms in the Jacobians[3] | Proposed (BA-LINS) |
|---|---|---|---|---|---|---|---|---|
| | $\sigma_\varepsilon = 0.01m$ | $\sigma_\varepsilon = 0.02m$ | $\sigma_\varepsilon = 0.05m$ | The second | The middle | The furthest | | |
| *campus* | 0.44 / 1.50 | **0.32 / 1.10** | 0.43 / 1.62 | 0.47 / 1.65 | 0.76 / 2.47 | 1.01 / 2.92 | 0.40 / 1.26 | 0.40 / 1.24 |
| *building* | 0.66 / 2.00 | 0.58 / 1.89 | 1.06 / 2.32 | 0.66 / 1.86 | 0.81 / 2.03 | 1.04 / 2.54 | **0.56 / 1.82** | 0.57 / **1.82** |
| *playground* | failed | 1.09 / 1.31 | 0.97 / 1.58 | 1.03 / 1.09 | 0.87 / 1.10 | 0.92 / 1.28 | 0.89 / 1.02 | **0.84 / 0.96** |
| *park* | failed | 1.01 / 1.70 | 1.52 / 2.62 | **0.75 / 1.69** | 0.98 / 1.72 | 1.07 / 1.74 | 1.22 / 2.07 | 1.23 / 2.07 |
| *cs_campus* | 0.50 / 1.48 | 0.47 / 1.01 | 1.45 / 2.64 | **0.36 / 1.20** | 0.38 / 1.31 | 0.48 / 1.52 | 0.43 / 1.20 | 0.55 / 1.39 |
| *luojia_square* | **0.38 / 1.51** | 0.63 / 2.36 | 1.28 / 6.21 | 0.45 / 2.06 | 0.49 / 2.52 | 1.04 / 5.61 | 0.55 / 2.82 | 0.54 / 2.72 |
| *east_lake* | 0.86 / 3.74 | 1.34 / 6.51 | 1.60 / 7.22 | **0.82 / 3.22** | 0.84 / 4.23 | 0.96 / 4.39 | 0.84 / 3.52 | 0.85 / 3.57 |
| *library* | 1.07 / 2.93 | **0.41 / 1.88** | 0.55/ **1.88** | 0.84 / 2.49 | 0.68 / 2.13 | 1.04 / 3.01 | 0.50 / 1.92 | 0.49 / 1.90 |
| Average | Invalid | 0.73 / 2.22 | 1.11 / 3.26 | **0.67 / 1.91** | 0.73 / 2.19 | 0.95 / 2.88 | **0.67** / 1.95 | 0.68 / 1.96 |

[1] $\sigma_\varepsilon$ represents the STD of the point-to-plane distance meaurement, and it can be converted to the covariance of the plane-point BA measurement using (7).
[2] Five points (the green and purple points in Fig. 4) are found in the keyframe point-cloud map, and they are reordered by the distance relative to the projected point (the red point in Fig. 4).
[3] These tiny terms are derived from the errors of the plane parameters $(n^w, d^w)$ caused by the IMU pose and the LiDAR-IMU extrinsic errors.

*F. The Impact of the Tiny Terms in Jacobians*

In Section IV.D.2, the errors of the plane parameters $(n^w, d^w)$ caused by the IMU pose errors and the LiDAR-IMU extrinsic errors are not considered in (11) and (13), due to their negligible impact. We also conduct experiments to evaluate the impact of these tiny terms in Jacobians. According to the results in Table VI, the AREs and ATEs are almost the same as those of BA-LINS when considering these tiny terms in the Jacobians. Consequently, the proposed method is reasonable and can also reduce computational complexity.

## VI. CONCLUSION

This paper presents an F2F BA for LiDAR-inertial navigation to improve the DR capability. We associate the same-plane points across multiple frames by building upon the F2F data association. Thus, an F2F plane-point BA measurement model is proposed to construct a multi-state constraint with an adaptive covariance estimation algorithm. The LiDAR plane-point BA and IMU-preintegration measurements are tightly coupled in a sliding-window optimizer. We conduct comprehensive real-world experiments on both public and private datasets. The results indicate that the proposed BA-LINS exhibits superior accuracy to SOTA LINSs. More specifically, the absolute translation accuracy is improved by 29.5% on the private dataset compared to the baseline system FF-LINS. Besides, the state-estimation efficiency is also improved by 28.7% due to the proposed plane-point BA measurement model. The ablation experiment results demonstrate that the proposed methods are reasonable and effective in improving the accuracy and efficiency.

However, current implementations in BA-LINS are mainly designed for Livox solid-state LiDARs. This is because the point-cloud coverage can be notably increased by accumulating multiple frames due to the non-repetitive scanning pattern of solid-state LiDARs. Only then can the proposed same-plane point association method and the plane-point BA measurement model be effective. Hence, the proposed method should apply to spinning LiDARs with more laser beams, such as 64 and 128. However, additional work may be necessary to apply the proposed BA method to 16-beam or 32-beam spinning LiDARs.

REFERENCES

[1] J. Lin and F. Zhang, "Loam livox: A fast, robust, high-precision LiDAR odometry and mapping package for LiDARs of small FoV," in *2020 IEEE International Conference on Robotics and Automation (ICRA)*, May 2020, pp. 3126–3131.
[2] X. Niu, H. Tang, T. Zhang, J. Fan, and J. Liu, "IC-GVINS: A Robust, Real-Time, INS-Centric GNSS-Visual-Inertial Navigation System," *IEEE Robotics and Automation Letters*, vol. 8, no. 1, pp. 216–223, Jan. 2023.
[3] P. D. Groves, *Principles of GNSS, inertial, and multisensor integrated navigation systems*. Boston: Artech House, 2008.
[4] H. Tang, T. Zhang, X. Niu, L. Wang, L. Wei, and J. Liu, "FF-LINS: A Consistent Frame-to-Frame Solid-State-LiDAR-Inertial State Estimator,"

*IEEE Robotics and Automation Letters*, vol. 8, no. 12, pp. 8525–8532, Dec. 2023.
[5] T. Shan, B. Englot, D. Meyers, W. Wang, C. Ratti, and D. Rus, "LIO-SAM: Tightly-coupled Lidar Inertial Odometry via Smoothing and Mapping," in *2020 IEEE/RSJ International Conference on Intelligent Robots and Systems (IROS)*, 2020, pp. 5135–5142.
[6] W. Xu and F. Zhang, "FAST-LIO: A Fast, Robust LiDAR-Inertial Odometry Package by Tightly-Coupled Iterated Kalman Filter," *IEEE Robotics and Automation Letters*, vol. 6, no. 2, pp. 3317–3324, Apr. 2021.
[7] W. Xu, Y. Cai, D. He, J. Lin, and F. Zhang, "FAST-LIO2: Fast Direct LiDAR-Inertial Odometry," *IEEE Transactions on Robotics*, pp. 1–21, 2022.
[8] J. Zhang and S. Singh, "Low-drift and real-time lidar odometry and mapping," *Auton Robot*, vol. 41, no. 2, pp. 401–416, Feb. 2017.
[9] C. Cadena *et al.*, "Past, Present, and Future of Simultaneous Localization and Mapping: Toward the Robust-Perception Age," *IEEE Trans. Robot.*, vol. 32, no. 6, pp. 1309–1332, Dec. 2016.
[10] C. Qin, H. Ye, C. E. Pranata, J. Han, S. Zhang, and M. Liu, "LINS: A Lidar-Inertial State Estimator for Robust and Efficient Navigation," in *2020 IEEE International Conference on Robotics and Automation (ICRA)*, May 2020, pp. 8899–8906.
[11] C. Bai, T. Xiao, Y. Chen, H. Wang, F. Zhang, and X. Gao, "Faster-LIO: Lightweight Tightly Coupled Lidar-Inertial Odometry Using Parallel Sparse Incremental Voxels," *IEEE Robotics and Automation Letters*, vol. 7, no. 2, pp. 4861–4868, Apr. 2022.
[12] G. Huang, "Visual-Inertial Navigation: A Concise Review," in *2019 International Conference on Robotics and Automation (ICRA)*, May 2019, pp. 9572–9582.
[13] T.-M. Nguyen, M. Cao, S. Yuan, Y. Lyu, T. H. Nguyen, and L. Xie, "VIRAL-Fusion: A Visual-Inertial-Ranging-Lidar Sensor Fusion Approach," *IEEE Transactions on Robotics*, vol. 38, no. 2, pp. 958–977, Apr. 2022.
[14] P. Geneva, K. Eckenhoff, Y. Yang, and G. Huang, "LIPS: LiDAR-Inertial 3D Plane SLAM," in *2018 IEEE/RSJ International Conference on Intelligent Robots and Systems (IROS)*, Madrid: IEEE, Oct. 2018, pp. 123–130.
[15] X. Zuo *et al.*, "LIC-Fusion 2.0: LiDAR-Inertial-Camera Odometry with Sliding-Window Plane-Feature Tracking," in *2020 IEEE/RSJ International Conference on Intelligent Robots and Systems (IROS)*, 2020, pp. 5112–5119.
[16] D. Wisth, M. Camurri, S. Das, and M. Fallon, "Unified Multi-Modal Landmark Tracking for Tightly Coupled Lidar-Visual-Inertial Odometry," *IEEE Robotics and Automation Letters*, vol. 6, no. 2, pp. 1004–1011, Apr. 2021.
[17] H. Ye, Y. Chen, and M. Liu, "Tightly Coupled 3D Lidar Inertial Odometry and Mapping," in *2019 International Conference on Robotics and Automation (ICRA)*, Montreal, QC, Canada: IEEE, May 2019, pp. 3144–3150.
[18] R. B. Rusu and S. Cousins, "3D is here: Point Cloud Library (PCL)," in *2011 IEEE International Conference on Robotics and Automation*, May 2011, pp. 1–4.
[19] H. Tang, X. Niu, T. Zhang, L. Wang, and J. Liu, "LE-VINS: A Robust Solid-State-LiDAR-Enhanced Visual-Inertial Navigation System for Low-Speed Robots," *IEEE Trans. Instrum. Meas.*, vol. 72, pp. 1–13, 2023.
[20] R. Hartley and A. Zisserman, *Multiple View Geometry in Computer Vision, Second Edition*. Cambridge university press, 2003.
[21] Z. Liu and F. Zhang, "BALM: Bundle Adjustment for Lidar Mapping," *IEEE Robotics and Automation Letters*, vol. 6, no. 2, pp. 3184–3191, Apr. 2021.
[22] W. Hess, D. Kohler, H. Rapp, and D. Andor, "Real-time loop closure in 2D LIDAR SLAM," in *2016 IEEE International Conference on Robotics and Automation (ICRA)*, May 2016, pp. 1271–1278.
[23] Z. Wang, L. Zhang, Y. Shen, and Y. Zhou, "D-LIOM: Tightly-Coupled Direct LiDAR-Inertial Odometry and Mapping," *IEEE Trans. Multimedia*, vol. 25, pp. 3905–3920, 2023.
[24] T. D. Barfoot, *State Estimation for Robotics*. Cambridge: Cambridge University Press, 2017.
[25] K. Li, M. Li, and U. D. Hanebeck, "Towards High-Performance Solid-State-LiDAR-Inertial Odometry and Mapping," *IEEE Robotics and Automation Letters*, vol. 6, no. 3, pp. 5167–5174, Jul. 2021.
[26] X. Zuo, P. Geneva, W. Lee, Y. Liu, and G. Huang, "LIC-Fusion: LiDAR-Inertial-Camera Odometry," in *2019 IEEE/RSJ International Conference on Intelligent Robots and Systems (IROS)*, Macau, China: IEEE, Nov. 2019, pp. 5848–5854.
[27] R. Szeliski, *Computer vision: algorithms and applications*. Springer Nature, 2022.
[28] Jianbo Shi and Tomasi, "Good features to track," in *1994 Proceedings of IEEE Conference on Computer Vision and Pattern Recognition*, Jun. 1994, pp. 593–600.
[29] E. Rublee, V. Rabaud, K. Konolige, and G. Bradski, "ORB: An efficient alternative to SIFT or SURF," in *2011 International Conference on Computer Vision*, Barcelona, Spain: IEEE, Nov. 2011, pp. 2564–2571.
[30] T. Qin, P. Li, and S. Shen, "VINS-Mono: A Robust and Versatile Monocular Visual-Inertial State Estimator," *IEEE Trans. Robot.*, vol. 34, no. 4, pp. 1004–1020, Aug. 2018.
[31] C. Campos, R. Elvira, J. J. G. Rodríguez, J. M. M. Montiel, and J. D. Tardós, "ORB-SLAM3: An Accurate Open-Source Library for Visual, Visual–Inertial, and Multimap SLAM," *IEEE Transactions on Robotics*, vol. 37, no. 6, pp. 1874–1890, 2021.
[32] Z. Liu, X. Liu, and F. Zhang, "Efficient and Consistent Bundle Adjustment on Lidar Point Clouds," *IEEE Trans. Robot.*, vol. 39, no. 6, pp. 4366–4386, Dec. 2023.
[33] X. Liu, Z. Liu, F. Kong, and F. Zhang, "Large-Scale LiDAR Consistent Mapping Using Hierarchical LiDAR Bundle Adjustment," *IEEE Robotics and Automation Letters*, vol. 8, no. 3, pp. 1523–1530, Mar. 2023.
[34] R. Li, X. Zhang, S. Zhang, J. Yuan, H. Liu, and S. Wu, "BA-LIOM: tightly coupled laser-inertial odometry and mapping with bundle adjustment," *Robotica*, pp. 1–17, 2024.
[35] A. I. Mourikis and S. I. Roumeliotis, "A multi-state constraint Kalman filter for vision-aided inertial navigation," in *Proceedings 2007 IEEE International Conference on Robotics and Automation*, IEEE, 2007, pp. 3565–3572.
[36] M. Li and A. I. Mourikis, "High-precision, consistent EKF-based visual-inertial odometry," *The International Journal of Robotics Research*, vol. 32, no. 6, pp. 690–711, May 2013.
[37] P. Geneva, K. Eckenhoff, W. Lee, Y. Yang, and G. Huang, "OpenVINS: A Research Platform for Visual-Inertial Estimation," in *2020 IEEE International Conference on Robotics and Automation (ICRA)*, Paris, France: IEEE, May 2020, pp. 4666–4672.
[38] J. Sola, "Quaternion kinematics for the error-state Kalman filter," *arXiv preprint arXiv:1711.02508*, 2017.
[39] G. Sibley, L. Matthies, and G. Sukhatme, "Sliding window filter with application to planetary landing: Sibley et al.: Sliding Window Filter," *J. Field Robotics*, vol. 27, no. 5, pp. 587–608, Sep. 2010.
[40] J. Engel, V. Koltun, and D. Cremers, "Direct Sparse Odometry," *IEEE Trans. Pattern Anal. Mach. Intell.*, vol. 40, no. 3, pp. 611–625, Mar. 2018.
[41] H. Tang, T. Zhang, X. Niu, J. Fan, and J. Liu, "Impact of the Earth Rotation Compensation on MEMS-IMU Preintegration of Factor Graph Optimization," *IEEE Sensors Journal*, vol. 22, no. 17, pp. 17194–17204, Sep. 2022.
[42] Agarwal, Sameer, Mierle, Keir, and The Ceres Solver Team, "Ceres Solver." Mar. 2022. [Online]. Available: https://github.com/ceres-solver/ceres-solver
[43] M. Li and A. I. Mourikis, "Online temporal calibration for camera–IMU systems: Theory and algorithms," *The International Journal of Robotics Research*, vol. 33, no. 7, pp. 947–964, 2014.
[44] J. Lin and F. Zhang, "R3LIVE: A Robust, Real-time, RGB-colored, LiDAR-Inertial-Visual tightly-coupled state Estimation and mapping package," in *2022 International Conference on Robotics and Automation (ICRA)*, 2022, pp. 10672–10678.
[45] J. Li *et al.*, "WHU-Helmet: A Helmet-Based Multisensor SLAM Dataset for the Evaluation of Real-Time 3-D Mapping in Large-Scale GNSS-Denied Environments," *IEEE Transactions on Geoscience and Remote Sensing*, vol. 61, pp. 1–16, 2023.
[46] Z. Zhang and D. Scaramuzza, "A Tutorial on Quantitative Trajectory Evaluation for Visual(-Inertial) Odometry," in *2018 IEEE/RSJ International Conference on Intelligent Robots and Systems (IROS)*, Madrid: IEEE, Oct. 2018, pp. 7244–7251.
[47] M. Grupp, "evo." Mar. 2023. [Online]. Available: https://github.com/MichaelGrupp/evo

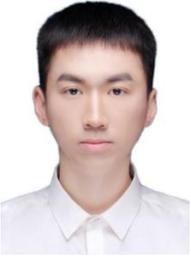
**Hailiang Tang** received the M.E. and Ph.D. degrees from Wuhan University, China, in 2020 and 2023, respectively. He is currently a postdoctoral fellow at the GNSS Research Center, Wuhan University. His research interests include autonomous robotics systems, visual and LiDAR SLAM, GNSS/INS integration, and deep learning.

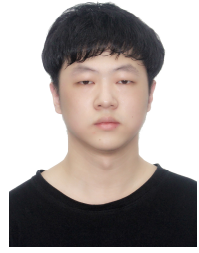
**Man Yuan** received the B.E. (with honors) degree in communication engineering from Wuhan University, Wuhan, China, in 2023, where he is currently pursuing a master's degree in navigation, guidance, and control with the GNSS Research Center. His primary research interests include GNSS/INS integrations and LiDAR-based navigation.

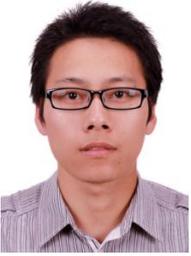
**Tisheng Zhang** is an associate professor in GNSS Research Center at Wuhan University, China. He holds a B.SC. and a Ph.D. in Communication and Information System from Wuhan University, Wuhan, China, in 2008 and 2013, respectively. From 2018 to 2019, he was a PostDoctor of the HongKong Polytechnic University. His research interests focus on the fields of GNSS receiver and multi-sensor deep integration.

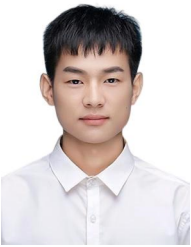
**Liqiang Wang** received the B.E. (with honors) and the M.E degrees from Wuhan University, China, in 2020 and 2023, respectively. He is currently pursuing a Ph.D. degree in GNSS Research Center, Wuhan University. His primary research interests include GNSS/INS integrations and visual-based navigation.

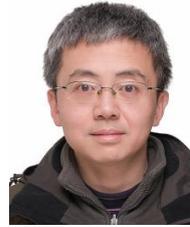
**Xiaoji Niu** received his bachelor's and Ph.D. degrees from the Department of Precision Instruments, Tsinghua University, in 1997 and 2002, respectively. He did post-doctoral research with the University of Calgary and worked as a Senior Scientist in SiRF Technology Inc. He is currently a Professor with the GNSS Research Center, Wuhan University, China. He has published more than 90 academic papers and own 28 patents. He leads a multi-sensor navigation group focusing on GNSS/INS integration, low-cost navigation sensor fusion, and its new applications.